\def\year{2016}\relax
\documentclass[letterpaper]{article}
\usepackage{aaai16}
\nocopyright

\usepackage{times}
\usepackage{helvet}
\usepackage{courier}
\usepackage{amsmath}
\usepackage{amssymb}
\usepackage{xcolor}
\usepackage{amsthm}
\usepackage[utf8]{inputenc}

\theoremstyle{definition}
\newtheorem{definition}{Definition}

\begin{document}
%
\title{Theory reconstruction: a representation learning view on predicate invention}
\author{Sebastijan Duman\v{c}i\'{c}, Wannes Meert, Hendrik Blockeel\\
Computer Science Department\\
KU Leuven, Belgium}
\maketitle

\section{Motivation}

Predicate invention is one of the core challenges of Inductive Logic Programming (ILP) and Statistical relational learning (SRL) since their beginnings.
The task is to extend the initial vocabulary that is given to a relational learner by discovering novel concepts and relations from data.
The discovered concepts should be explained in terms of the observable ones.
The invented predicates can also be \textit{statistical} if the uncertainty in the discovered predicates is represented explicitly.

The benefits of predicate invention are numerous.
Firstly, it can produce more compact and comprehensive models by capturing dependencies between observed predicates, which consequently yields less parameters and reduces the risk of overfitting.
Secondly, as each invented predicate can later be re-used, it allows a learner to take larger steps through the search space.
Finally, invented predicates can represent latent states of a data-generating process, potentially increasing the performance of a model. 

The progress so far, however, has been limited.
We argue here that one of the main obstacles is the lack of a framework that formalizes the problem.
The existing work is a collection of individual approaches harnessing different ideas, greedily searching for new predicates to improve classification accuracy or invent predicates to compress a complete logical program (known as theory revision).

In this work, we propose a unifying framework for statistical predicate invention and theory revision.
Our proposal departs from the conventional approaches and addresses it from the perspective of \textit{unsupervised representation learning} \cite{Bengio2009}.
The main motivation for this proposal lies in the key ingredient of representation learning success: representation learning methods proved to be very effective at constructing many layers of features, that can be re-used to address the final classification task. 
This significantly resembles the idea behind predicate invention.

We argue here that the construction of layers of features can be seen as \textit{propositional predicate invention}, where each hidden node, a new feature, can be seen as a binary variable dependent on the states of a subset of variables in the preceding layer.
Furthermore, discovered hidden variables can later compose more complex dependencies throughout the layers of a deep model.
Secondly, we argue that relational learners suffer from conceptually the same problem as the tasks successfully addressed by deep learning: that of high dimensionality.
The connection comes from the interpretation that formulas in a relational model are seen as boolean features of the model.
Given a knowledge base and its vocabulary, the search space of possible formulas (or features) is huge even in domains with a small number of predicates, and is therefore high-dimensional.
When learning the structure, learners aim at selecting a small subset of all possible formulas that are most relevant for the task.

We base the framework on a particular line of research within deep learning: \textit{autoencoders} \cite{Vincent2010,NIPS2006_3048,Vincent2008,Tieleman2008} and \textit{sparse coding} methods \cite{NIPS2006_2979,GregorL10}.
These approaches take a \textit{generative view} on representation learning. They create a \textit{hidden representation} able to re-generate the original data from a smaller subset of features.
Autoencoders achieve this by means of a neural network with a single hidden layer and putting input, instead of the labels, at the output of the same neural network.
Sparse coding approaches, on the contrary, discover a set of hidden vectors that would, by linear combination, reconstruct the original examples.
Both approaches are instantiations of the \textit{encoder-decoder} approach, where one learns an \textit{encoder} to map original data to a hidden representation, and a separate \textit{decoder} to reconstruct the original data from the hidden representation.
Both methods can be further \textit{stacked} to obtain layers of features.
Deep models built in this manner are proven to be effective in extracting useful features in a completely unsupervised way, successfully applied to text and image recognition.

With this proposal, we intend to contribute towards bridging the relational and deep learning communities on the problem of predicate invention.
The main underlying idea is to encode the provided set of features into a new set of \textit{latent features} that could reconstruct the majority of the original features. 

\section{State of the Art}
\label{sec:RelWork}

Within the ILP research, predicates can be invented by analyzing first-order formulas, and forming a predicate to represent either their commonalities \cite{Wogulis1989} or their differences \cite{MuggletonBu88}.
A weakness of such approaches is that they are prone to over-generating predicates, many not useful ones.
Predicates can also be invented by instantiating second-order templates \cite{Silverstein91}, or to represent exceptions to learned rules \cite{Srinivasan92}.
More recently, \citeauthor{MuggletonMIL} (\citeyear{MuggletonMIL}) introduce a \textit{meta-interpreter} perspective on predicate invention.

Within the SRL research, \citeauthor{Popescul2004} (\citeyear{Popescul2004}) apply k-means clustering to the objects of each type in a domain, create predicates to represent clusters, and learn relations among them.
\citeauthor{Perlich2003} (\citeyear{Perlich2003}) present a number of approaches for aggregating multi-relational data.
\citeauthor{Craven2001} (\citeyear{Craven2001}) propose a learning mechanism for hypertext domains in which class predictions produced by naive Bayes are added to an ILP system (FOIL) as invented predicates.
\citeauthor{DavisOSBPC07} (\citeyear{DavisOSBPC07}) learn Horn clauses with an off-the-shelf ILP system, create a predicate for each clause learned, and add it as a feature to the database.
\citeauthor{Kok07statisticalpredicate} (\citeyear{Kok07statisticalpredicate}) cluster both predicate and constant symbols to create new predicates.
\citeauthor{WangMC15} (\citeyear{WangMC15}) capture differences between similar formulas and represent it with a new predicate.

\section{New view: Theory reconstruction}

Our proposal is based on the \textit{encoder-decoder} architecture, utilized by many representation learning approaches.
An essential difference between our approach and previously proposed ones is that the auto-encoder does not try to encode the knowledge base, but only the patterns that occur in it.
Furthermore, in contrast to the majority of the previous approaches, our approach is entirely unsupervised.


Let $\mathcal{KB}$ be a knowledge base  with a vocabulary $ \mathcal{V} = \{\mathcal{C}, \mathcal{P}\}$, with $\mathcal{P}$ a set of predicates and $\mathcal{C}$ a set of constants. 
A \textit{sentence} in the first-order logic is a formula with no free variables.
Let $\mathcal{L}$ be a set of sentences called the language bias (typically specified using syntactic constraints, e.g., all conjunctive formulas containing at most 3 literals and at most 2, existentially quantified, variables).
Given a knowledge base $\mathcal{KB}$, let $\mathcal{T}$ be a set of all sentences in $\mathcal{L}$ that are true in $\mathcal{KB}$.

As enumeration of all possible satisfiable formulas w.r.t. $\mathcal{KB}$ would be infeasible, $\mathcal{L}$ represents a trade-off between expressivity and efficiency.
Therefore, $\mathcal{L}$ plays an important role that significantly influences the kind of predicates that will be invented.

\noindent \textbf{Example 1.} \textit{Assume $\mathcal{P}$$=$\{{\footnotesize \textit{smokes/1, cancer/1, friends/2}}\}, $\mathcal{C}$$=$\{{\footnotesize \textit{jane, john}}\} and $\mathcal{KB}$$=$\{{\footnotesize smokes(john), cancer(john), friends(john,jane), smokes(jane)}\}. 
Let $\mathcal{L}$ be all existentially quantified conjunctive formulas with connected variables with each term being a variable, and length in range (2,3).
Then $\mathcal{T}$ is \{{\scriptsize smokes(X),cancer(X); smokes(X),friends(X,Y);  smokes(Y),friends(X,Y); cancer(X),friends(X,Y); smokes(X),friends(X,Y),smokes(Y); cancer(X),friends(X,Y),smokes(Y); smokes(X),cancer(X),friends(X,Y) }\}.}

Let $\mathcal{Q}$ be a set of predicates that do not occur in $\mathcal{P}$; these predicates are called \textit{hidden predicates}.
Let $\mathcal{F}$ be a set of sentences with the following properties: $\mathcal{F}$ contains exactly one definition for each predicate in $\mathcal{Q}$, and each definition is of the form $h(X_1, \ldots, X_k) \Leftrightarrow Body$ where $Body$ is a sentence built using $\mathcal{P}$ and $h$ is a predicate symbol from $\mathcal{Q}$.
Let $\mathcal{KB}_{\mathcal{Q}}$ contain truth assignments to $p \in \mathcal{Q}$ given their corresponding definitions and constants $\mathcal{C}$.

\begin{definition}{\textbf{Hidden representation.}}
Let $\mathcal{L}_{\mathcal{Q}}$ be a language bias with a vocabulary $\mathcal{Q}$.
A \textit{hidden representation} $\mathcal{T}_{\mathcal{Q}}$ of $\mathcal{T}$ is a set of sentences in $\mathcal{L}_{\mathcal{Q}}$ that are true in $\mathcal{KB}_{\mathcal{Q}}$.
\end{definition}

\noindent \textbf{Example 2.} \textit{Assume $\mathcal{Q}=$ \{{\footnotesize $h_1/1, h_2/2, h_3/2, h_4/2$}\} and $\mathcal{F}=$ \{{\scriptsize $h_1(X)\Leftrightarrow smokes(X) \land cancer(X)$, $h_2(X,Y)\Leftrightarrow smokes(X) \land friends(X,Y)$, $h_3(X,Y)\Leftrightarrow cancer(X) \land friends(X,Y)$, $h_4(X,Y) \Leftrightarrow smokes(Y) \land friends(X,Y)$} \}.
Limiting $\mathcal{L}_{\mathcal{Q}}$ to existentially quantified conjunctive formulas of maximal length 2, $\mathcal{T}_{\mathcal{Q}}$ is \{{\scriptsize $h_1$(X); $h_2$(X); $h_4$(X,Y); $h_3$(X,Y); $h_2$(X,Y),$h_4$(X,Y); $h_3$(X,Y),$h_4$(X,Y); $h_2$(X,Y),$h_1$(X); $h_2$(X,Y),$h_2$(X,Y); $h_1$(X,Y),$h_2$(X,Y); $h_1$(X,Y),$h_3$(X,Y)}\}}.

\begin{definition}{\textbf{Relational auto-encoder.}}
A relational auto-encoder is a program that, given a logical theory $\mathcal{T}$, constructs an encoder $\mathcal{E}$ and a decoder $\mathcal{D}$, together with the $(\mathcal{Q}, \mathcal{F})$.
\end{definition}

\begin{definition}{\textbf{Relational encoder.}}
A relational encoder is a function $\mathcal{E}: 2^L \rightarrow 2^{\mathcal{L}_{\mathcal{Q}}}$ that maps a set of sentences in $\mathcal{L}$ to a set of sentences in $\mathcal{L}_{\mathcal{Q}}$, given $(\mathcal{Q}, \mathcal{F})$.
\end{definition}

\begin{definition}{\textbf{Relational decoder.}}
A relational decoder  is a function $\mathcal{D}: 2^{\mathcal{L}_{\mathcal{Q}}} \rightarrow 2^L$ that maps a hidden representation to a set of sentences in $\mathcal{L}$.
\end{definition}

\begin{definition}{\textbf{Theory reconstruction.}} 
The theory reconstruction task is then defined as learning the $(\mathcal{Q},\mathcal{F})$ such that
\begin{align*}
	& \underset{(\mathcal{Q},\mathcal{F})}{\text{minimize}} & & \left( \mathcal{T} \circleddash \mathcal{D}(\mathcal{E}(\mathcal{T})) \right) - q(\mathcal{Q}, \mathcal{F}, \mathcal{KB}) 	
\label{eq:KBR}
\end{align*}
where $\circleddash$ is a difference measure between two logical theories, and $q(\mathcal{Q}, \mathcal{F}, \mathcal{KB})$ measures a \textit{quality} of the hidden representation according to a specified criterion, such as sparsity, compression or others.
\end{definition}

The main role of $q$ is to prevent the identity mapping between $\mathcal{P}$ and $\mathcal{Q}$ which, though lossless, would be a useless one.
In contrast to the neural auto-encoder, where one fixes the structure and learns just the weights, the proposed framework learns the structure itself while putting equal weights on all connections.
One can further instantiate the $(\mathcal{Q},\mathcal{F})$ representation and repeat the procedure to obtain more complex predicates.

\noindent \textbf{Example 3.} \textit{
Assuming $\mathcal{T}$ from Ex.~1 and $\mathcal{Q},\mathcal{F}$ and $\mathcal{T}_{\mathcal{Q}}$ in Ex.~2,
$\mathcal{D}$ can then perfectly reconstruct $\mathcal{T}$ using the subset of $\mathcal{T}_{\mathcal{Q}}$ and substituting predicates in $\mathcal{Q}$ with their definitions: \{{\scriptsize $h_1$(X); $h_2$(X); $h_4$(X,Y); $h_3$(X,Y); $h_2$(X,Y),$h_4$(X,Y); $h_3$(X,Y),$h_4$(X,Y); $h_2$(X,Y),$h_1$(X)}\} with the same order as $\mathcal{T}$ in Ex.~1. 
In contrast to $\mathcal{T}$, the representation using hidden predicates is more concise since it requires less atoms per formula to represent exactly the same knowledge. 
In practice one hopes that $\mathcal{KB}_{\mathcal{Q}}$ would as well contain less facts than the original $\mathcal{KB}$.
}

This formulation encapsulates both predicate invention and theory revision, in which case $\mathcal{T}$ is substituted with the formulas in an existing model $M$, $\mathcal{T}_M$.
Moreover, the formulation can be further extended to account on uncertainty by means of \textit{weighted reconstruction}, where weights can resemble a probability of a formula being true in data, similar to many SRL approaches. 

This formulation intends to establish a common ground between relational and deep learning, and start a discussion to define a framework for predicate invention.

\newpage

\bibliographystyle{aaai}
\bibliography{starai}

\end{document}